\documentclass[11pt]{article}

\usepackage[final]{acl}

\usepackage{times}
\usepackage{latexsym}

\usepackage[T1]{fontenc}

\usepackage[utf8]{inputenc}

\usepackage{microtype}

\usepackage{inconsolata}

\usepackage{graphicx}

\usepackage{amsmath}

\usepackage{multirow}
\usepackage{booktabs}

\newcommand{\TP}[1]{$\mathcal{TP}_{#1}$}
\newcommand{\IP}[1]{$\mathcal{IP}_{#1}$}

\usepackage{array}  
\usepackage{tabularx} 
\usepackage{booktabs}  
\usepackage{ragged2e} 

\usepackage{float} 
\usepackage{arydshln}

\usepackage{multirow}  

\usepackage{amsmath}

\usepackage{arydshln} 

\newcommand{\mydashline}{%
  \noalign{\vskip0.5ex}%
  \multispan{11}\hbox to \hsize{\leaders\hbox to 2.5pt{\hss-\hss}\hfil}%
  \noalign{\vskip0.5ex}%
}

\usepackage{booktabs} 

\usepackage{enumitem}

\usepackage{CJKutf8}

\usepackage{subdepth} 

\usepackage{makecell} 
\newcolumntype{N}{>{\centering\arraybackslash}p{1.2cm}} 
\newcolumntype{F}{>{\centering\arraybackslash}p{0.8cm}} 

\usepackage{tabularx}
\usepackage{multirow}
\usepackage{booktabs}
\usepackage[table]{xcolor}
\usepackage{geometry}
\title{Seeing the Poem: Image-Semantic Detection of AI-Generated Modern Chinese Poetry with MLLMs}

\author{
       Shanshan Wang$^1$~~~~
       Fengying Ye$^1$~~~~
       Hanjia Lyu$^2$~~~~
       Caiwen Gou$^3$~~~~
       Junchao Wu$^1$~~~~\\
       \textbf{Jingming Yao}$^4$~~~~
       \textbf{Chengzhong Xu}$^1$~~~~
       \textbf{Jiebo Luo}$^{2,*}$~~~~
       \textbf{Derek F. Wong}$^{1,*}$~~~~
       \\
    $^1$Department of Computer and Information Science, University of Macau \\
     $^2$University of Rochester, $^3$Sichuan University\\
     $^4$Department of Portuguese, Faculty of Arts and Humanities, University of Macau\\ 
\texttt{nlp2ct.shanshan@gmail.com}, \texttt{jluo@cs.rochester.edu}, \texttt{derekfw@um.edu.mo} \\\\
    }

\begin{document}
\maketitle

\begingroup
\def\thefootnote{\relax}\footnotetext{$^*$Corresponding Author}
\endgroup

\begin{abstract}

Previous detection studies have shown that LLMs cannot be effectively used as detectors, but these studies have not addressed modern Chinese poetry. Moreover, no relevant research has explored the performance of LLMs in detecting modern Chinese poetry. This paper evaluates and enhances the performance of LLMs as detectors for modern Chinese poetry, and proposes an image-semantic guided poetry detection method. Compared with traditional detection approaches, our method innovatively incorporates images that reflect the content of the poetry. Through example-driven approaches, our method effectively integrates information such as meaning, imagery, and feeling from the image, then forms a complementary judgment with the poem text. Experimental results demonstrate that the LLM detectors based on our method outperform baseline detectors based on plain text, and even surpass the best-performing traditional detector, RoBERTa. The Gemini detector using our method achieves a Macro-F1 score of 85.65\%, reaching the state-of-the-art level. The performance improvements of different LLM detectors on multiple LLMs-generated data prove the effectiveness of our method.

\end{abstract}

\section{Introduction}

The rapid advancement of Large Language Models (LLMs) has significantly improved the quality of AI-generated text, making it increasingly similar to human-written content~\cite{lai2024adaptive}. While this progress demonstrates notable technical achievements, it has also raised concerns about the growing difficulty in distinguishing AI-generated text from genuine human-written text~\cite{jakesch2023human,porter2024ai,hayawi2024imitation,najjar2025leveraging}.

\begin{figure}[ht!]
    \centering
    \includegraphics[width=0.9\linewidth]{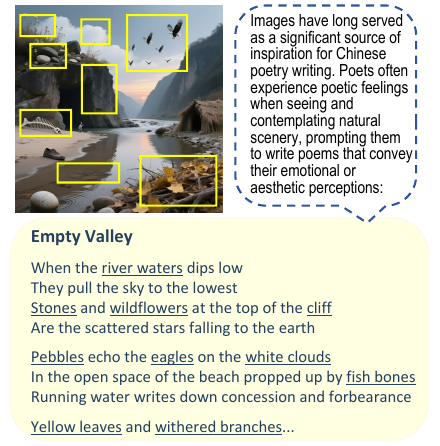}
    \caption{The relationship between human poetry and images. In this example, key nouns from the poem are underlined, and their corresponding visual elements are highlighted with yellow bounding boxes in the image.}
    \label{fig:poem_example}
\end{figure}



The blurring boundary between human and AI-generated text now poses a particularly pressing challenge, extending even into the domain of poetry~\cite{wang-etal-2025-benchmarking}. Poetry, as a precious part of human cultural heritage, carries deep historical and cultural significance, permeating various aspects of our lives comprehensively~\cite{ren2023generating, zhang2019through}. With a history spanning thousands of years, it has profoundly influenced the development of different nations and cultures~\cite{yi2018chinese, zhao2022poetrybert}. Alarmingly, the total amount of AI-generated poetry has been predicted to exceed all human-composed poetry throughout history~\cite{huojunming2020}.


\begin{figure*}[t]
    \centering
\includegraphics[width=0.9\textwidth]{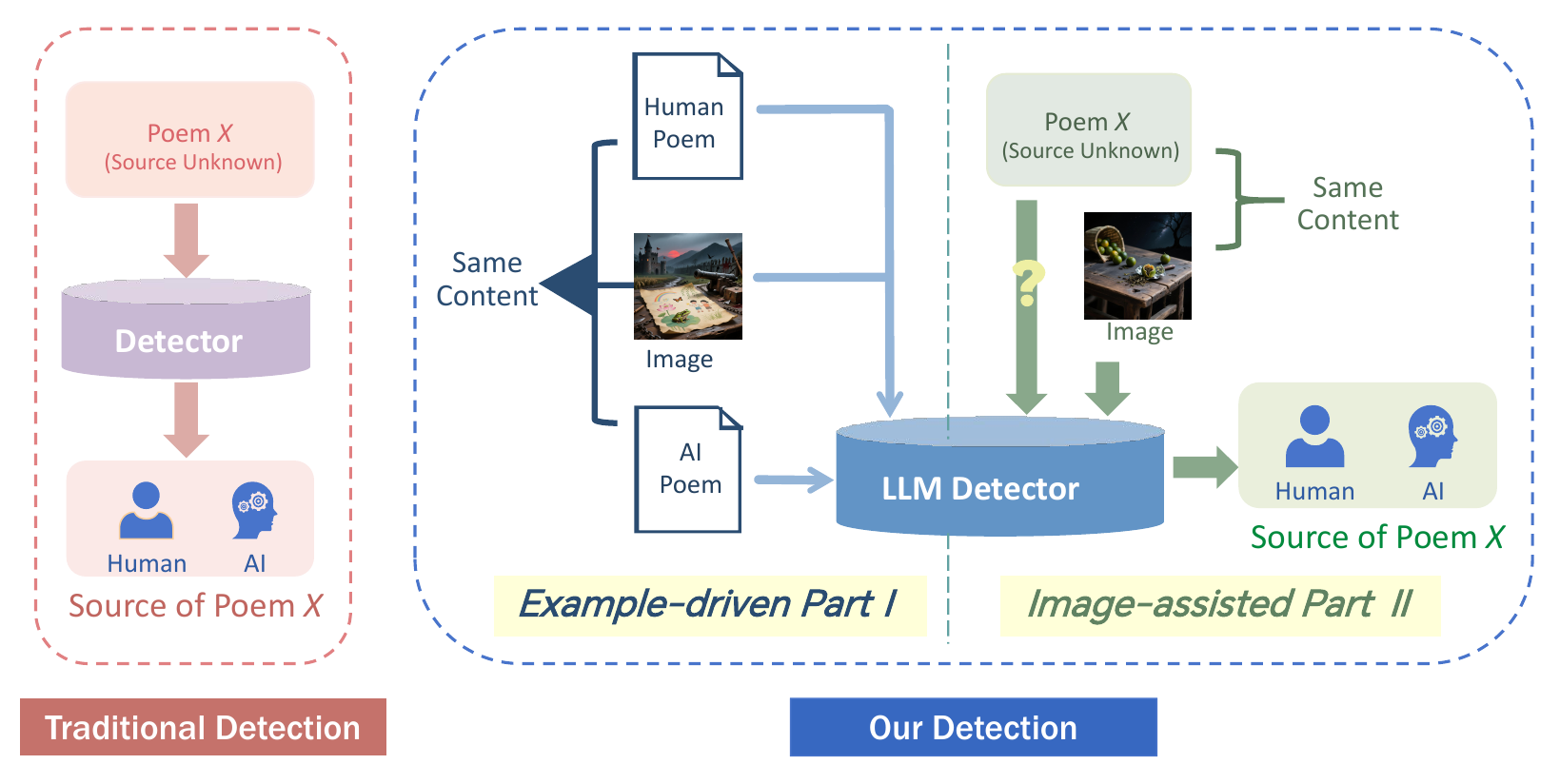}
    \caption{Comparison between the framework of the traditional detection method and our proposed IMAGINE method.}
    \label{framework}
\end{figure*}

In China, since the release of DeepSeek-R1 \cite{guo2025deepseek}, the quantity of LLMs-generated poetry has grown rapidly. Some individuals now use LLMs to imitate the style of famous poets to generate modern Chinese poems and submit them to regular literary journals. This phenomenon creates considerable confusion and disruption for editors, readers, and even professional poets, as humans often struggle to distinguish AI-generated poems from those written by humans~\cite{kobis2021artificial,jakesch2023human, darda2023value}. Therefore, the development of effective automatic detection methods has become not only necessary but also urgent~\cite{wu2025survey}. 




LLMs have been applied to various tasks and have demonstrated varying degrees of performance \cite{lin-etal-2025-large, ye2025unveiling, LyuBias,  shen2025measuring, lan-etal-2025-f2bench, lan2025mcbe, shen2026preconditioned}, making them promising as detectors for AI-generated text. Previous research in the NLP field on modern Chinese poetry was limited, mainly focusing on translation \cite{song2023towards, wang2024best} and comprehension \cite{wang-etal-2026-chatgpt}. A recent benchmark study~\cite{wang-etal-2025-benchmarking} has revealed that traditional text detectors such as Fast-DetectGPT~\cite{bao2023fast}, Binoculars~\cite{hans2024spotting}, LRR~\cite{su2023detectllm}, Log-Rank~\cite{gehrmann2019gltr}, Log-Likelihood~\cite{solaiman2019release}, and RoBERTa-based classifiers~\cite{RoBERTa2019}, which perform well in general text domains, are \textbf{\emph{not}} reliable for detecting AI-generated modern Chinese poetry. In fact, most detectors other than RoBERTa achieve less than 70\% accuracy in detecting LLM-generated poems. 



 In the real world, poetry often originates from the poet's interaction with visual scenes, a principle summarized in the famous concept of ``expressing emotions through scenery.'' In other words, the poet sees the image first, and then creates the poem.  Specifically, poetry is the artistic presentation of subjective emotions inspired by objective external objects~\cite{HuMin2003}. Images have long served as a significant source of inspiration for Chinese poetry writing~\cite{cheng2018image, li2021poetic}. Expressing emotions through scenery is a common method in poetry writing~\cite{WangDeming2003, XiongZhengrong2001}. Poets often experience poetic feelings when seeing and contemplating natural scenery~\cite{li2021poetic}, prompting them to write poems that express their emotional or aesthetic perceptions~\cite{xu2018images, cheng2018image}. However, the essence of poetry lies not in describing the objective facts in an image, but in condensing the objects, scenes, and emotions within them to reveal the deeper meaning and poetic essence behind them \cite{liu2018beyond}.  Images, in this context, carry rich information, extending beyond immediate visual details to encompass higher-level semantic themes~\cite{xu2018images}.  Take Figure~\ref{fig:poem_example} as an example, imagery often emerges across scattered lines and through seemingly simple nouns such as river, cliff, eagles, fish bones, and yellow leaves. They collectively form a coherent emotional landscape or artistic conception. This emotional core is rarely stated explicitly, but rather constructed through the subtle interplay of visuals, themes, and affect.

Building on this poetically grounded generation setting where visual and emotional imagery plays a central role,  rather than relying solely on textual analysis, models can benefit from visual semantics to better grasp the implicit meanings and emotional depth encoded in poetry.
So we propose a novel image-semantic guided poetry detection framework based on Multi-modal Large Language Models
(MLLMs) , termed \textbf{IMAGINE}, which integrates multimodal information to further refine LLM-based detection. 


Specifically, we first provide the model with paired examples of AI and human poems along with corresponding images reflecting the content of the poem. 
The model then receives both the poem under examination and a related image to guide its classification. This framework leverages image-semantic guidance to deepen the model's understanding of poetic nuance and origin.



Experimental results demonstrate that our image-semantic guided method yields substantial improvements in poetry detection task.
Gemini, when enhanced with our approach, achieves a state-of-the-art Macro-F1 score of 85.65\% across all LLM-generated and human-written poems. Our findings also reveal intriguing trends: different LLMs vary in their detection performance, suggesting that not all are equally suitable for this task. 
Furthermore, different categories of detectors exhibit varying performance when distinguishing between LLM-generated and human-written poems. Notably, our method significantly narrows the performance gap within LLM-based detectors. 
The consistent improvements observed across both overall and fine-grained metrics demonstrate the robustness and effectiveness of our image-semantic guided detection framework.

Main contributions of our work are as follows:

\begin{itemize}[itemsep=0pt]
       
    \item We construct a dataset for detecting AI-generated modern Chinese poetry that is grounded in poetic creation practices, comprising 800 human-written poems and 3,200 LLM-generated poems. Complementary to prior datasets, our dataset requires LLMs to retain comprehensive features of human poetry, including theme, meaning, imagery, and emotion/feeling, thus enabling a semantically grounded evaluation setting.


    \item We propose IMAGINE, a novel image-semantic guided detection framework, which achieves state-of-the-art performance on poetry detection and surpassing all baselines.

    \item We determine the initial performance of LLMs as a poetry detector and effectively enhance their detection capability through an example-driven approach.





\end{itemize}

\section{Related Work}

Previous research has achieved notable progress in detecting LLM-generated text \cite{chen2025repreguard}, but these approaches face limitations when applied to modern Chinese poetry. \citet{wu2024wrote} proposed a detection method based on grammatical error correction scores. This method leverages differences in grammatical errors as a discriminative feature to distinguish between texts originating from LLMs and those authored by humans, achieving effective results across various real-world text sources. However, this method is not applicable to modern Chinese poetry because modern poetry does not require adherence to grammatical rules \cite{blohm2018sentence, ChenZhongyi2012grammar, DengCheng2007Dilemma}. \citet{kobis2021artificial} attempted a novel Turing test by increasing judges' motivation to improve the accuracy of their judgments. They assessed human ability to distinguish rhymed, rhythmic English poems generated by GPT-2, including human preferences for model-generated and human-written poems. However, the results demonstrated that humans could not accurately identify AI-generated poems. \citet{nguyen2023detect} proposed a machine learning classification method based on handcrafted feature engineering and a feature detection method based on text similarity to detect human-written and synthetically generated content. The first method extracts multi-dimensional features such as lexical, syntactic, semantic, and readability features, achieving an F1 score as high as 99.93\% using models such as Random Forest (RF) \cite{kam1995random} and XGBoost (XGB) \cite{chen2016xgboost}. The second method, through inverse generation and cosine similarity calculation, can effectively distinguish between human-generated and ChatGPT-generated text without requiring real text. Recently, \citet{li2026wrote} proposed a benchmark for detecting classical Chinese poetry and evaluated the performance of various AI detectors. Their results demonstrate that current Chinese text detectors are ineffective in detecting classical Chinese poetry generated by LLMs.

\begin{table}[t]
\centering
\begingroup
\small
\begin{tabular}{lcccc}
\toprule 
&  {Poems} & {Stanzas} & {Lines} & {Words} \\
\midrule 
Human &  800	& 2.1K	& 15.0K	& 137.5K \\
\midrule
GPT-4.1  &  800	& 2.3K	& 15.5K	& 173.5K\\
GLM-4  & 800	& 2.2K	& 15.1K	& 132.7K \\
DeepSeek-V3 & 800	& 2.8K	& 15.3K	& 119.3K \\
DeepSeek-R1 &  800	& 3.1K	& 16.6K	& 172.2K \\
\bottomrule 
\end{tabular}
\endgroup
\caption{\label{poems data}
The statistics of our poem data.
}
\end{table}

\section{Dataset}
\label{Dataset}

Our dataset consists of 800 human-written poems, 3,200 LLMs-generated poems, and 800 images corresponding to the human-written poems.

\paragraph{Human-written Poems} The human-written poems employed in our work are sourced directly from the AIGenPoetry dataset~\cite{wang-etal-2025-benchmarking}, which comprises 800 modern Chinese poems authored by professional poets. These poems feature well-documented provenance and exhibit diversity in style, form, and theme, ensuring both high quality and representativeness for our study.

\paragraph{LLMs-generated Poems} We follow \citet{wang-etal-2025-benchmarking} and employ four LLMs including GPT-4.1 (OpenAI)~\cite{achiam2023gpt}, GLM-4 (Zhipu AI)~\cite{zhipu2024}, DeepSeek-V3~\cite{liu2024deepseek}, and DeepSeek-R1~\cite{guo2025deepseek} to generate one corresponding poem for each human-written poem.  For poem generation, we use the recommended hyperparameters from DeepSeek-R1~\cite{liu2024deepseek}: temperature set to \texttt{1.5} and top\_p to \texttt{0.95}. This results in a total of 3,200 LLMs-generated poems. The specific prompt used is provided in Table~\ref{Poem Generation Prompt}. Table~\ref{poems data} presents the summary statistics of our poem dataset.

\begin{table}[t]
\centering
\small
\begin{tabular}{p{7.2cm}}
\hline
Please create a new modern Chinese poem. It must share the same title as the poem provided below, and its theme, meaning, imagery, emotion/feeling, etc., must be completely identical to the original. Furthermore, it must retain all the nouns from the provided poem but be composed in a completely different writing style. The output should contain only the poem itself, without any parsing, analysis, or indication of poetic genre. \\
Poem: \textit{\{\{$P_i$\}\}}   \\
\hline
\end{tabular}
\normalsize 
\renewcommand{\arraystretch}{1.0}
\caption{\label{Poem Generation Prompt}The prompt for poem generation.
}
\end{table}

\paragraph{Corresponding Images} 
To support image-semantic guided detection, we associate each human-written poem with a representative image. We first select 30 poems from the dataset, covering diverse styles. Next, six mainstream image generation models are prompted according to our designed prompt (shown in Table~\ref{Image Generation Prompt}) to generate one corresponding image per poem. Thus, each model generates a total of 30 images. These models include DALL-E 3~\cite{betker2023improving}, GPT-Image-1~\cite{Gpt-image-1}, GPT-4o-image~\cite{hurst2024gpt}, Doubao, Qwen-image~\cite{wu2025qwen}, and Hidream~\cite{hidreami1technicalreport}. 


To determine the best model for large-scale generation, we invite professional poets to evaluate the images produced by different models. Based on their assessments, we select Qwen-Image as the image generation model and prompt it to generate one image for each of the 800  human-written poems. 




\section{Text-Only Detection}\label{sec:motivation_analysis}

In this section, we aim to empirically demonstrate the limitations of text-only detectors in reliably distinguishing between human-written and LLM-generated modern Chinese poetry.


\begin{table}[t]
\centering
\begingroup
\small
\begin{tabular}{lccccc}
\toprule 
\textbf{Gen. $\rightarrow$}&  {GPT} & {GLM} & {V3.1} & {R1} & {Avg.}\\
\midrule
\textbf{Fast-Detec.} &  33.33& 	51.96& 	62.73& 	33.28	& 45.33
 \\
\textbf{LRR}  & 33.83	& 39.31	& 48.18	& 33.33	& 38.66
 \\
\textbf{Log-Likeli.} &  33.61	& 45.03	& 48.93	& 33.33	& 40.23
 \\
\textbf{Log-Rank}  & 33.61	& 43.56	& 47.17	& 33.33	& 39.42
 \\
\textbf{Binoculars} &   34.32	& 61.73	& 69.08	& 33.83	& 49.74
\\
\textbf{RoBERTa}  &  89.44	& 73.31	& 78.44	& 96.74	& 84.48
\\
\addlinespace[0.1em] 
\cdashline{2-6}
\addlinespace[0.4em] 
\textbf{Avg.} & 43.02	& 52.48	& 59.09	& 43.97	& 49.64
\\
\bottomrule 
\end{tabular}
\endgroup
\caption{\label{detectors}
The performance (Macro-F1, \%) of various traditional detectors. Det. and Gen. represent Detector and Generator respectively. GPT, GLM, V3.1, and R1 represent GPT-4.1, GLM-4, DeepSeek-V3.1, and DeepSeek-R1, respectively. In the first column, Fast-Detet. and Log-Likeli. represent Fast-DetectGPT and Log-Likelihood, respectively.
}
\end{table}




\subsection{Traditional Text-Only Detection}


Following \citet{wang-etal-2025-benchmarking}, we evaluate a total of six traditional detectors, including Fast-DetectGPT~\cite{bao2023fast}, LRR~\cite{su2023detectllm}, Log-Likelihood~\cite{solaiman2019release}, Log-Rank~\cite{gehrmann2019gltr}, Binoculars~\cite{hans2024spotting}, and RoBERTa-based classifier~\cite{RoBERTa2019}.
The implementation details are provided in Appendix~\ref{appendix:implementation_traditional}.

We apply these detectors to the dateset described in Table~\ref{poems data}. Both the testing and training sets consist of 400 pairs of poems, with each pair containing one LLM-generated poem and one human-written poem. Table~\ref{detectors} presents the performance (Macro F1-scores) of traditional detectors while Tables~\ref{RoBERTa} through \ref{Binoculars} provide detailed results of different detectors, respectively, including the Macro-F1 and F1 scores for both AI-generated (F1\_AI) and human-written poems (F1\_Human). 


Evidently, 
traditional detectors remain incapable of reliably identifying LLMs-generated poems. Even the best-performing detector, RoBERTa, achieves a Macro-F1 score of only 73.31\% for GLM-4-generated poems.


\subsection{LLMs-based Text Detection}
\label{LLMs-based Text Detection}

We further evaluate the performance of LLM-based text-only detectors under two prompting strategies: vanilla prompting and example-driven prompting.


\paragraph{Initial Detection}


To evaluate the initial ability of LLMs in distinguishing between human-written and different LLMs-generated poetry, we design a baseline prompt, denoted as \TP{1}. The complete \TP{1} is presented in Table~\ref{TP1}. Subsequently, mainstream LLMs including qwen-vl-max-2025-08-13, GPT-5, and gemini-3-flash-preview are prompted to judge whether the poem texts in the dataset are created by AI or humans. Table \ref{text macro-f1} presents the Macro-F1 scores. We also report the F1-scores for AI poems as positive examples (denoted as F1\_AI), and the F1-scores for human poems as positive examples (denoted as F1\_Human), as shown in the third row of Tables~\ref{Qwen text}, \ref{GPT-5 text}, and \ref{Gemini text}.

The results show that, except for Gemini, which scores higher than RoBERTa in Macro-F1 (74.92\%) for poems generated by GLM-4, other LLM detectors score lower than RoBERTa in F1 for poems generated by different LLMs. This demonstrates that LLM-based plain-text detectors are ineffective at distinguishing between AI-generated poems and human-written poems.

\begin{table}[t]
\centering
\begingroup
\small
\begin{tabular}{lccccc}
\toprule 
\textbf{Gen. $\rightarrow$}&  {GPT} & {GLM} & {V3.1} & {R1} & {Avg.}\\
\midrule
\textbf{Detector $\rightarrow$}  & \multicolumn{5}{c}{\textbf{Gemini}} \\ 
\cmidrule(lr){2-6}
\TP{1}  &  81.58	& 74.92	& 69.70	& 54.90&  70.28\\
\TP{2}  & 86.86	& 77.46	& 69.31	& 84.35 & 79.50 \\
\midrule
\textbf{Det. $\rightarrow$}  & \multicolumn{5}{c}{\textbf{Qwen}} \\ 
\cmidrule(lr){2-6}
\TP{1} &  33.44	& 34.80	& 33.16	& 33.44	& 	33.71 \\
\TP{2}  &  44.15	& 47.55	& 33.73	& 42.64	& 	42.02 \\
\midrule
\textbf{Detector $\rightarrow$}  & \multicolumn{5}{c}{\textbf{GPT-5}} \\ 
\cmidrule(lr){2-6}
\TP{1} & 74.33	&69.23	&46.07	&48.08 &  59.43\\
\TP{2}   & 74.75	&70.70	&49.54	&76.89 & 67.97\\
\bottomrule 
\end{tabular}
\endgroup
\caption{\label{text macro-f1}
The performance (Macro-F1, \%) of LLM-based plain-text detectors (Gemini, Qwen, GPT-5) in distinguishing between poems generated by different LLMs and human-written poems.
}
\end{table}

\paragraph{Enhanced Detection}

We enhance the detection capability of LLMs for poetry texts through an example-driven approach. Specifically, we design a new plain text prompt, denoted as \TP{2}, as shown in Table~\ref{TP2}. Based on the vanilla baseline prompt \TP{1}, we add a pair of examples containing AI-generated and human-written poems to \TP{2}, specifying the relationship between the two poems in the examples that their titles, themes, meanings, imagery, and moods/emotions are identical. Next, different LLMs are prompted to determine whether the source of the poem to be judged was AI or human. Table~\ref{text macro-f1} presents the results (Macro-F1) of different LLM-based plain-text detector using \TP{2} to judge poems generated by different LLMs and human-written poems. Tables~\ref{Qwen text}, \ref{GPT-5 text}, and \ref{Gemini text} respectively present the detailed performance, including F1\_AI, F1\_Human, and Macro-F1.

Experimental results show that our designed example-driven method improves the overall Macro-F1 score to 79.50\% (Table~\ref{text macro-f1}). When employing \TP{2}, Qwen, GPT-5, and Gemini achieve varying degrees of improvement in Macro-F1 scores for poetry generated by LLMs other than DeepSeek-V3.1. Particularly for poems generated by DeepSeek-R1, Gemini using \TP{2} achieves a 29.45\% improvement in Macro-F1 compared to \TP{1} (Table~\ref{Gemini text}), while GPT-5 achieves a 28.81\% improvement (Table~\ref{GPT-5 text}). Similar to \TP{1}, Gemini (77.46\%) using \TP{2} also outperforms RoBERTa (73.31\%) in Macro-F1 scores for poems generated by GLM-4. However, LLM detectors still remain unreliable in judging poetry generated by LLMs other than GPT-4.1 and DeepSeek-R1.





\begin{table}[t]
\centering
\small
\begin{tabular}{p{7.2cm}}
\hline
You are a professional expert in analyzing modern Chinese poetry, skilled at distinguishing between human-written and AI-generated poems.
 \\
 \addlinespace[0.5em]
The following is an image and two modern Chinese poems, where the AI poem was generated by AI based on the human-written poem. The AI poem shares identical title, theme, meaning, imagery, mood/emotion, and other content with the human poem, and retains all the nouns from the human poem, differing only in its writing style. The image was also generated by AI in a realistic style based on the human poem, with particular attention paid to the poem's title, theme, meaning, imagery, and mood/emotion. It embodies all the noun elements present in the human poem. 
\\
\addlinespace[0.5em]
Image: \textit{\{\{Example of Image\}\}} \\
Human Poem: \textit{\{\{Example of Human Poem\}\}} \\
AI Poem: \textit{\{\{Example of AI Poem\}\}}
\\
\addlinespace[0.5em]
The following is a new image and a new modern Chinese poem generated by AI in a realistic style based on a human poem, with particular attention paid to the poem's title, theme, meaning, imagery, and emotion/feeling, and it embodies all the noun elements from that human poem. Now, please determine whether the following modern Chinese poem was written by human or generated by AI. Please note that the object of your judgment is the ``Poem Text'', not the image. Output only the final judgment as either ``Human'' or ``AI.'' The output format should be: 
Answer: [Human or AI] 
\\
\addlinespace[0.5em]
Image: \textit{\{\{Image corresponding to the Poem Text to be judged\}\}} \\
Poem Text: \textit{\{\{Poem text to be judged\}\}} \\
\hline
\end{tabular}
\normalsize 
\renewcommand{\arraystretch}{1.0}
\caption{\label{IP3}The image-semantic guided prompt (\IP{3}) we designed for poetry detection.
}
\end{table}

\begin{table}[t]
\centering
\begingroup
\small
\begin{tabular}{lccccc}
\toprule 
\textbf{Gen. $\rightarrow$}&  {GPT} & {GLM} & {V3.1} & {R1} & {Avg.}\\
\midrule
\textbf{Det. $\rightarrow$}  & \multicolumn{5}{c}{\textbf{Gemini}} \\ 
\cmidrule(lr){2-6}
\IP{2} &  88.61	& 81.08	& 68.99	& 83.86	& 	80.64 \\
\IP{3}  & \textbf{ 91.87}	& \textbf{81.48}	& 74.51	& 94.75	& 	\textbf{85.65} \\
\midrule
\textbf{Det. $\rightarrow$}  & \multicolumn{5}{c}{\textbf{Qwen}} \\ 
\cmidrule(lr){2-6}
\IP{3} &  63.25	& 61.98	& 46.46	& 74.96	& 	61.66 \\
\bottomrule 
\end{tabular}
\endgroup
\caption{\label{image macro-f1}
The performance (Macro-F1, \%) of LLM-based detectors (Gemini, Qwen) applying our proposed method in distinguishing between poems generated by different LLMs and human-written poems.
}
\end{table}

\begin{table*}[h]
\centering
\small
\begin{tabularx}{\textwidth}{l X}
\toprule
\textbf{Prompts} & \textbf{Characteristics} \\
\midrule
\TP{1} & The model is required to determine whether the poem to be judged originates from AI or human. \TP{1} is used as the baseline.\\
\addlinespace[0.5em] 
\TP{2}  & 
1) The model is first \textbf{provided with a pair of examples} containing one AI-generated poem and one human-written poem, along with the \textbf{relationship between the two poems} in the example. 2) Then, the model is required to determine whether the poem to be judged originates from AI or human.\\
\midrule
\IP{2} & 
Based on \TP{2}. 1) The model is first \textbf{provided with a pair of examples} containing one AI-generated poem and one human-written poem, \textbf{an image} reflecting the content of the poems in the example, the \textbf{relationship between the image and the two poems}, and the \textbf{relationship between the two poems}.
2) Then, the model is provided with \textbf{an image reflecting the content of the poem to be judged}, and the \textbf{relationship between this image and this poem}. Subsequently, the model is required to determine whether the poem to be judged originates from AI or human.\\
\addlinespace[0.5em]
\IP{3} & 
Based on \TP{2}. 1) The model is first \textbf{provided with a pair of examples} containing one AI-generated poem and one human-written poem, \textbf{an image} reflecting the content of the poems in the example, the \textbf{relationship between the image and the two poems}, and the \textbf{relationship between the two poems}. Furthermore, the model is informed of \textbf{\textit{the detailed source of image and each poem in the example}}. For example, the image is generated by AI in a realistic style based on human-written poem. 2) Then, the model is provided with \textbf{an image that reflects the text content of the poem to be judged}, the \textbf{relationship between this image and this poem}, and \textit{\textbf{the detailed source of the image}}. Subsequently, the model is required to determine whether the poem to be judged originated from AI or human. \\
\bottomrule
\end{tabularx}
\caption{\label{Characteristics of Detection Prompt}The focuses and characteristics of different prompts we designed for poetry detection.
}
\end{table*}

\section{Image-Semantic Guided Poetry Detection}


Images are the significant source of inspiration for Chinese poetry writing~\cite{cheng2018image, li2021poetic}. Expressing emotion and feeling through scenery is a common method in poetry writing~\cite{WangDeming2003, XiongZhengrong2001}. Furthermore, images carry extremely rich information, transcending mere visual details and even delving into semantic themes~\cite{xu2018images}.


To more accurately determine the source of poetry, we propose a novel image-semantic guided poetry detection method named \textbf{IMAGINE} (IMAge-semaNtic GuIded poEtry detection). Compared to text-only poetry detection, IMAGINE innovatively introduces images that reflect the content of the poem text. IMAGINE consists of two parts: 

\textbf{}

\begin{itemize}[itemsep=0pt]
       
    \item In the first part, we provide the model with a pair of examples comprising an AI-generated poem and a human-written poem, along with an image reflecting the content of the poems in the examples. We also inform the model of the relationship between the image and the two poems, the relationship between the two poems themselves, and the specific source of the image and each poem in the examples. 

    \item In the second part, we provide the model with an image reflecting the content of the poem to be judged, informing it of the relationship between the image and this poem, and the specific source of the image. The model is then asked to determine whether the poem to be judged originates from AI or human.

\end{itemize}

Both the AI-generated poem and the image are generated from the same human-written poem. Figure~\ref{framework} presents the framework of our proposed method. The complete prompt (\IP{3}) of IMAGINE is presented in Table~\ref{IP3}. Based on \TP{2}, we also design another image-semantic guided prompt, denoted as \IP{2}, which is presented in Table~\ref{IP2} of the Appendix~\ref{subsec:Prompts Designed for Poetry Detection}. The different focuses and characteristics of all prompts are presented in Table~\ref{Characteristics of Detection Prompt}. The AI-generated poem share the same title, theme, meaning, imagery, and emotion/feeling as the human poem. At the beginning of each prompt, the model is asked to act as a professional expert in analyzing modern Chinese poetry.



\section{Experiment}

In Section~\ref{LLMs-based Text Detection}, we show that Gemini is the best-performing model for distinguishing the source of poem text, while the performance of Qwen is less satisfactory. Therefore, in subsequent main experiments, we employ Gemini as the primary detector and Qwen as the secondary detector to implement our proposed IMAGINE method. Specifically, Gemini is utilized to determine the source of poems generated by different LLMs and human written poems through \IP{2} and \IP{3}, while Qwen is used to judge the source of poems through \IP{3}.


Table~\ref {image macro-f1} presents the performance (Macro-F1) of LLM-based detectors (Gemini, Qwen, GPT-5) applying our proposed method in distinguishing between poems generated by different LLMs and human-written poems. Tables~\ref{Gemini image} and \ref{Qwen image} present the detailed performance (\%) of LLM-based detector (Gemini and Qwen), including F1\_AI, F1\_Human, and Macro-F1.


\begin{table*}
\centering
\small
\begin{tabularx}{\textwidth}{l *{6}{>{\centering\arraybackslash}X}}
\toprule
\textbf{F1-score $\rightarrow$} & \textbf{F1\_AI} & \textbf{F1\_Human} & \textbf{Macro-F1} & \textbf{F1\_AI} & \textbf{F1\_Human} & \textbf{Macro-F1} \\
\cmidrule(lr){2-4} \cmidrule(lr){5-7}  
\textbf{Generator $\rightarrow$}  & \multicolumn{3}{c}{\textbf{GPT-4.1}} & 
 \multicolumn{3}{c}{\textbf{GLM-4}}\\
\cmidrule(lr){2-4} \cmidrule(lr){5-7} 
\IP{2}  & 89.00	& 88.23	& 88.61	& 80.21	& 81.96	& 81.08
 \\
\IP{3}  & 91.68	& 92.06	& \textbf{91.87}	& 78.77	& 84.19	& \textbf{81.48}
 \\
\midrule
\textbf{Generator $\rightarrow$}  & \multicolumn{3}{c}{\textbf{DeepSeek-V3.1}} & 
\multicolumn{3}{c}{\textbf{DeepSeek-R1}}\\
\cmidrule(lr){2-4} \cmidrule(lr){5-7} 
\IP{2}  & 64.53	& 73.44	& 68.99	& 83.44	& 84.29	& 83.86
 \\
\IP{3}  &  69.47	& 79.54	& 74.51	& 94.59	& 94.90	& \textbf{94.75}
 \\
\bottomrule
\end{tabularx}
\caption{\label{Gemini image} The detailed performance (\%) of LLM-based detector (Gemini) applying our proposed method on poems generated by different LLMs and human-written poems, including F1\_AI, F1\_Human, and Macro-F1.  
}
\end{table*}

\section{Analysis}

\subsection{Effectiveness of Our Method}

Our proposed IMAGINE method (\IP{3}) effectively improves the detection performance of LLMs for poetry by integrating images reflecting the content of the poems.

\paragraph{Gemini Detector} Table~\ref{image macro-f1} shows that the overall performance of the LLM detector based on our method (\IP{3}) is superior to the baseline LLM detector based on plain text (\TP{1}) and the enhanced plain text detector (\TP{2}), and even surpasses the best-performing traditional detector, RoBERTa. Specifically, the Gemini detector using our method (\IP{3}) achieves an overall Macro-F1 score of 85.65\% for all poems generated by LLMs and poems written by humans, reaching the state-of-the-art level. This represents a 15.37\% improvement over the baseline (\TP{1}, 70.28\%) and a 6.15\% improvement over the enhanced plain text detector (\TP{2}, 79.50\%) under the same settings but without added images. Furthermore, it is noteworthy that the overall Macro-F1 score of our method (\IP{3}) is also superior to RoBERTa (84.48\%) by 1.17\%, and surpasses all other detectors (Binoculars) except RoBERTa by at least 35.91\%. These improvements demonstrate that the rich information contained in the images is beneficial for the model's judgment of poetic text, strongly proving the effectiveness of our proposed IMAGINE method. Similarly, Gemini also achieves an overall improvement in detection performance compared to the baseline (\TP{1}) after using \IP{2} with added images corresponding to the poems, with the overall Macro-F1 increasing from 70.28\% to 80.64\%.


For individual LLM-generated poems and human-written poems, the Gemini detector using our method also achieves surprising results. For example, Gemini as a detector achieves a Macro-F1 of 94.75\% for detecting DeepSeek-R1-generated poems and human-written poems, exceeding the baseline (\TP{1}, 54.90\%) by 39.85\%. Furthermore, Gemini using \IP{3} outperforms RoBERTa in detecting poems generated by GPT-4.1 and GLM-4, as well as human-generated poems. Especially for the judgment of GLM-4 and human-generated poems, Macro-F1 score of \IP{3} (81.48\%) is 8.17\% higher than RoBERTa.

\paragraph{Qwen Detector} While Qwen's detection performance for LLM-generated and human-written poems remains unreliable, with an average Macro-F1 score of only 61.66\%, its overall detection performance and its performance for individual LLM-generated and human-written poems are significantly better than traditional text-based (\TP{1}) methods (Table~\ref{text macro-f1}).

The best-performing (Gemini) and worst-performing (Qwen) models based on text detection methods both show varying degrees of performance improvement after using our proposed IMAGINE method. This again highlights the effectiveness of images in poetry detection, suggesting that deeper information beyond visual details plays a crucial role. Using our method, the model effectively integrates information such as theme, meaning, imagery, and emotion/feeling from the image, forming complementary judgments with the text. For example, nouns like river, sunset, fallen leaves, thatch, and worn-out shoes in a poem might be distributed across different lines, and the poet uses the imagery represented by these nouns to construct a artistic conception. This artistic conception itself contains the poet's feelings and emotions, but is not explicitly presented in the text. However, when these nouns or images appear together in a single scene or image, they naturally become a unified whole and are more visually impactful, evoking emotional responses or deep reflection. This is the most important reason why we require AI-generated poetry and images to retain or reflect the nouns found in human poetry. 

\subsection{Meaningful Findings}

\paragraph{Performance Differences Among LLM Detectors.}

Different LLMs exhibit varying detection performance for poetry. Table~\ref{text macro-f1} shows that Qwen and GPT-5 have limited initial detection capabilities, especially when processing poetry from DeepSeek-V3.1 and DeepSeek-R1. This indicates that not all LLMs are inherently capable of reliable poetry text identification, and therefore not all LLMs are suitable for use as poetry detectors.


\paragraph{Detection Differences between AI and Human Poetry.}

Different types of detectors exhibit differences in their performance when dealing with poems generated by LLMs and human-written poems. Specifically, traditional detectors, including RoBERTa (Table~\ref{RoBERTa}), LRR (Table~\ref{LRR}), Log-Likelihood (Table~\ref{Log-Likelihood}), Log-Rank (Table~\ref{Log-Rank}), and Binoculars (Table~\ref{Binoculars}), are better at identifying poems generated by LLMs because their average F1\_AI score is higher than their F1\_Human score. Conversely, Fast-DetectGPT (Table~\ref{Fast-DetectGPT}) is better at detecting human-written poems. However, there are extreme cases: Fast-DetectGPT scores 0 on F1\_AI for poems generated by GPT-4.1 and DeepSeek-R1, while LRR, Log-Likelihoo, and Log-Rank score  0 on F1\_AI for poems generated by DeepSeek-R1, but achieve an F1\_Human score of 66.66\%. This indicates that traditional detectors, except for RoBERTa, tend to classify poems generated by DeepSeek-R1 as human when they cannot identify them. The misclassification of a large number of LLM poems as human poems reflects the severe inadequacy of traditional detectors in capturing the features of AI-generated poetry.

Unlike traditional detectors, LLMs exhibit completely opposite performance. As shown in Tables~\ref{Gemini text}, \ref{Qwen text}, and \ref{GPT-5 text}, Gemini, Qwen, and GPT-5 are all better at identifying human-written poems, with significantly higher F1\_Human scores than F1\_AI scores (\TP{1}). However, there are still extreme cases where Qwen's F1\_AI scores (\TP{1}) for poems generated by GPT-4.1, DeepSeek-V3, and DeepSeek-R1 are close to 0. This indicates that Qwen has no ability to distinguish poems generated by these LLMs under baseline settings.

However, promisingly, after applying our method (Table~\ref{Qwen image}, \IP{3}), Qwen's F1\_AI scores (\TP{1}) for poems generated by GPT-4.1, DeepSeek-V3, and DeepSeek-R1 improved to 63.43\%, 38.17\%, and 76.99\%, respectively. Furthermore, it is worth noting that the LLM detector using our method (\IP{3}) significantly reduces the gap between the F1\_AI score and the F1\_Human score compared to the baseline (\TP{1}).

The improvements in both overall and detailed performance across different LLM detectors repeatedly underscore the effectiveness of our method.

\section{Conclusion}

In this paper, we evaluate and enhance the performance of LLMs as detectors of modern Chinese poetry and propose an image-semantic guided poetry detection method based on LLMs. Compared with traditional detection methods, our method innovatively introduces images reflecting the content of the poems. Experimental results demonstrate that the LLMs detector based on our method outperforms plain text-based LLMs detectors and even surpasses the best-performing traditional detector, RoBERTa. The Gemini detector using our method achieves a comprehensive Macro-F1 score of 85.65\% for all LLMs-generated poems and human-written poems, reaching the state-of-the-art level. The improved performance of different LLMs detectors on different LLMs-generated data validates the effectiveness of our proposed method.

\section*{Limitations}

Our method draws inspiration from the way humans write poetry, where poets often express their emotions and thoughts by writing poems based on images or scenes they see in reality. Moreover, poetry typically contains numerous images represented by nouns. Therefore, incorporating images during detection can effectively enhance model performance. However, our method may not be fully applicable to other types of text, such as news articles or academic papers.



\section*{Ethics Statement}
 
The human poems used in this study are derived from existing datasets, and all authors of these poems have authorized our detection work to use the data.

\bibliography{main}

\appendix


\section{Appendix}
\label{sec: appendix}

\subsection{Prompt for Image Generation}
\label{subsec:Prompt for image generation}

Table~\ref{Image Generation Prompt} presents the prompt we designed for image generation.

\begin{table}[t]
\centering
\small
\begin{tabular}{p{7.2cm}}
\hline
Please generate a realistic-style image for the following modern Chinese poem, paying special attention to its title, theme, meaning, imagery, and emotion/feeling. The image must depict all the noun elements present in the poem. \\
\textbf{Important Note}: It is absolutely prohibited to include any form of text in the image—including Chinese characters, letters, numbers, symbols, or any text-like elements. Any textual content would severely degrade the image quality. You must avoid generating any text 100\%. \\
Poem: \textit{\{\{$P_i$\}\} }  \\
\hline
\end{tabular}
\normalsize 
\renewcommand{\arraystretch}{1.0}
\caption{\label{Image Generation Prompt}The prompt for image generation.
}
\end{table}

\subsection{The Role of Image}
\label{The Role of Image}

In the real world, poetry often originates from the poet's interaction with visual scenes, a principle summarized in the famous concept of ``expressing emotions through scenery.'' In other words, the poet sees the image first, and then creates the poem. This is the inspiration for our proposed method, whose design principles are as follows:

\paragraph{1) Image assists models in understanding implicit poetic meanings.} In our method, both the AI-generated poems and images originate from the same human-written poem, explicitly requiring the preservation of the poem's theme, meaning, imagery, and emotions. Specifically, first, as described in Section \ref{Dataset}, we require the AI-generated poems to retain all nouns from the human poem (Table~\ref{Poem Generation Prompt}), while the images must depict all noun elements present in the human poem (Table \ref{Image Generation Prompt}). Then, we invite professional poets to review the generated AI poems and images. This ensures that the images and AI-generated poems are semantically consistent with the same source (human poem).

Our experimental results and analysis show that the detector captures the deeper, more nuanced meanings between AI and human poetry with the aid of images.

\paragraph{2) Images provide a consistent reference.} All test instances in our work (including both human and AI poems) use images derived from the same human poem. In other words, the images don't provide any privileged information about authorship because they correspond identically to both human and AI poems. Therefore, our successful detection stems from the model's ability to integrate multimodal information to better understand the deeper meaning of the poems.

As we discussed in Section 7.1, nouns like ``river,'' ``sunset,'' and ``fallen leaves'' may be scattered across different lines of a poem, but when they appear together in the same image, they form a coherent visual scene that evokes emotional resonance. This allows the model to grasp the implicit emotions and deeper meanings of the poem, which are not explicitly expressed in the text. This deeper understanding derived from the image is key to improving detection performance.


\subsection{Implementation Details of Traditional Detectors}\label{appendix:implementation_traditional}
For traditional statistics-based detection methods, including Log-Likelihood, Log-Rank, and LRR, we utilize Qwen2.5-3B\footnote{\url{https://huggingface.co/Qwen/Qwen2.5-3B}} as the scoring model to extract logits-based features. For Fast-DetectGPT, we use Qwen2.5-3B as the reference model and Qwen2.5-7B\footnote{\url{https://huggingface.co/Qwen/Qwen2.5-7B}} as the scoring model. In the case of Binoculars, Qwen2.5-7B serves as the observer model, while Qwen2.5-7B-Instruct\footnote{\url{https://huggingface.co/Qwen/Qwen2.5-7B-Instruct}} acts as the performer. 

Regarding the RoBERTa-based classifier, we fine-tune a pre-trained Chinese RoBERTa model\footnote{\url{https://huggingface.co/hfl/chinese-RoBERTa-wwm-ext}}. 
The model is fine-tuned for \texttt{3} epochs with a batch size of \texttt{16} and a learning rate of \texttt{1e-6}.

\subsection{Prompts for Poetry Detection}
\label{subsec:Prompts Designed for Poetry Detection}

Tables \ref{TP1}, \ref{TP2}, and \ref{IP2} respectively present the different prompts we designed for poetry detection.

\begin{table}[t]
\centering
\small
\begin{tabular}{p{7.2cm}}
\hline
You are a professional expert in analyzing modern Chinese poetry, skilled at distinguishing between human-written and AI-generated poems. Now, please determine whether the following modern Chinese poem was written by human or generated by AI. Output only the final judgment as either ``Human'' or ``AI.'' The output format should be:  

Answer: [Human or AI] \\[0.5em]
Poem Text: \textit{\{\{Poem text to be judged\}\}}
 \\
\hline
\end{tabular}
\normalsize 
\renewcommand{\arraystretch}{1.0}
\caption{\label{TP1}The text-only prompt (\TP{1}) we designed for poetry detection.
}
\end{table}

\begin{table}[t]
\centering
\small
\begin{tabular}{p{7.2cm}}
\hline
You are a professional expert in analyzing modern Chinese poetry, skilled at distinguishing between human-written and AI-generated poems.\\[0.5em]

The following are two modern Chinese poems with identical title, theme, meaning, imagery, and emotion/feeling. One is written by human, and the other is generated by AI.
\\[0.5em]
Human Poem: \textit{\{\{Example of Human Poem\}\}}

AI Poem: \textit{\{\{Example of AI Poem\}\}}
\\[0.5em]
Now, please determine whether the following modern Chinese poem was written by human or generated by AI. Output only the final judgment as either ``Human'' or ``AI.'' The output format should be:  

Answer: [Human or AI] 

\\[0.5em]

Poem Text: \textit{\{\{Poem text to be judged\}\}}

 \\
\hline
\end{tabular}
\normalsize 
\renewcommand{\arraystretch}{1.0}
\caption{\label{TP2}The text-only prompt (\TP{2}) we designed for poetry detection.
}
\end{table}

\begin{table}[t]
\centering
\small
\begin{tabular}{p{7.2cm}}
\hline
You are a professional expert in analyzing modern Chinese poetry, skilled at distinguishing between human-written and AI-generated poems.
\\[0.5em]

The following is an image and two modern Chinese poems, with the image reflecting the majority of the textual content of both poems. The two poems share identical title, theme, meaning, imagery, emotion/feeling. One poem was written by human, and the other was generated by AI.
\\[0.5em]
Image: \textit{\{\{Example of Image\}\}}

Human Poem: \textit{\{\{Example of Human Poem\}\}}

AI Poem:\textit{ \{\{Example of AI Poem\}\}}
\\[0.5em]

The following is a new image and a new modern Chinese poem, with the image reflecting the majority of the textual content of this poems. Now, please determine whether the following modern Chinese poem was written by human or generated by AI. Please note that the object of your judgment is the ``Poem Text'', not the image. Output only the final judgment as either ``Human'' or ``AI.'' The output format should be:  

Answer: [Human or AI]  
\\[0.5em]
Image: \textit{\{\{Image corresponding to the Poem Text to be judged\}\}}

Poem Text: \textit{\{\{Poem text to be judged\}\}}
 \\
\hline
\end{tabular}
\normalsize 
\renewcommand{\arraystretch}{1.0}
\caption{\label{IP2}The image-semantic guided prompt (\IP{2}) we designed for poetry detection.
}
\end{table}



\subsection{Detailed Results}
\label{Detailed Results}

\paragraph{Traditional Detectors} Tables \ref{Fast-DetectGPT}, \ref{LRR}, \ref{Log-Likelihood}, \ref{Log-Rank}, and \ref{Binoculars} present the detailed performance of different traditional detectors, respectively.

\begin{table}[t]
\centering
\begingroup
\small
\begin{tabular}{lccc}
\toprule 
&  {F1\_AI} & {F1\_Human} & {Macro-F1}  \\
\midrule 
GPT-4.1	& 90.21	& 88.68	& 89.44\\
GLM-4	& 74.55	& 72.08	& 73.31\\
DeepSeek-V3.1	& 80.41	& 76.48	& 78.44\\
DeepSeek-R1	& 96.67	& 96.83	& 96.74 \\
\addlinespace[0.1em] 
\cdashline{2-4}
\addlinespace[0.4em] 
\textbf{Avg.} & 85.46	&83.52&	84.48
\\
\bottomrule 
\end{tabular}
\endgroup
\caption{\label{RoBERTa}
The detailed performance (\%) of RoBERTa on poems generated by different LLMs and human-written poems, including F1\_AI, F1\_Human, and Macro-F1.
}
\end{table}

\begin{table}[t]
\centering
\begingroup
\small
\begin{tabular}{lccc}
\toprule 
&  {F1\_AI} & {F1\_Human} & {Macro-F1}  \\
\midrule 
GPT-4.1	& 0.00	& 66.66		& 33.33\\
GLM-4	& 53.4		& 50.52		& 51.96\\
DeepSeek-V3.1	& 59.56		& 65.9		& 62.73
\\
DeepSeek-R1	& 0.00		& 66.56		& 33.28
\\
\addlinespace[0.1em] 
\cdashline{2-4}
\addlinespace[0.4em] 
\textbf{Avg.} &28.24		& 62.41		& 45.33
\\
\bottomrule 
\end{tabular}
\endgroup
\caption{\label{Fast-DetectGPT}
The detailed performance (\%) of Fast-DetectGPT on poems generated by different LLMs and human-written poems, including F1\_AI, F1\_Human, and Macro-F1.
}
\end{table}

\begin{table}[t]
\centering
\begingroup
\small
\begin{tabular}{lccc}
\toprule 
&  {F1\_AI} & {F1\_Human} & {Macro-F1}  \\
\midrule 
GPT-4.1	& 66.67		& 0.99		& 33.83
\\
GLM-4	& 66.09		& 12.53		& 39.31
\\
DeepSeek-V3.1	& 51.35		& 45.01		& 48.18
\\
DeepSeek-R1	& 0.00		& 66.66		& 33.33
\\
\addlinespace[0.1em] 
\cdashline{2-4}
\addlinespace[0.4em] 
\textbf{Avg.} &46.03		& 31.30 		& 38.66
\\
\bottomrule 
\end{tabular}
\endgroup
\caption{\label{LRR}
The detailed performance (\%) of LRR on poems generated by different LLMs and human-written poems.
}
\end{table}

\begin{table}[t]
\centering
\begingroup
\small
\begin{tabular}{lccc}
\toprule 
&  {F1\_AI} & {F1\_Human} & {Macro-F1}  \\
\midrule 
GPT-4.1	& 66.72		& 0.5		& 33.61
\\
GLM-4	& 64.95		& 25.1		& 45.03
\\
DeepSeek-V3.1	& 60.1		& 37.76		& 48.93
\\
DeepSeek-R1	& 0.00		& 66.66		& 33.33
\\
\addlinespace[0.1em] 
\cdashline{2-4}
\addlinespace[0.4em] 
\textbf{Avg.} &47.94		& 32.51		& 40.23
\\
\bottomrule 
\end{tabular}
\endgroup
\caption{\label{Log-Likelihood}
The detailed performance (\%) of Log-Likelihood on poems generated by different LLMs and human-written poems.
}
\end{table}

\begin{table}[t]
\centering
\begingroup
\small
\begin{tabular}{lccc}
\toprule 
&  {F1\_AI} & {F1\_Human} & {Macro-F1}  \\
\midrule 
GPT-4.1	& 66.72		& 0.5		& 33.61
\\
GLM-4	& 65.22		& 21.91		& 43.56
\\
DeepSeek-V3.1	& 60.18		& 34.16		& 47.17
\\
DeepSeek-R1	& 0.00		& 66.66		& 33.33
\\
\addlinespace[0.1em] 
\cdashline{2-4}
\addlinespace[0.4em] 
\textbf{Avg.} &48.03		& 30.81		& 39.42
\\
\bottomrule 
\end{tabular}
\endgroup
\caption{\label{Log-Rank}
The detailed performance (\%) of Log-Rank on poems generated by different LLMs and human-written poems.
}
\end{table}

\begin{table}[t]
\centering
\begingroup
\small
\begin{tabular}{lccc}
\toprule 
&  {F1\_AI} & {F1\_Human} & {Macro-F1}  \\
\midrule 
GPT-4.1	& 66.67		& 1.97		& 34.32
\\
GLM-4	& 65.61		& 57.86		& 61.73
\\
DeepSeek-V3.1	& 63.75		& 74.41		& 69.08
\\
DeepSeek-R1	& 66.67		& 0.99		& 33.83
\\
\addlinespace[0.1em] 
\cdashline{2-4}
\addlinespace[0.4em] 
\textbf{Avg.} &65.68		& 33.81		& 49.74
\\
\bottomrule 
\end{tabular}
\endgroup
\caption{\label{Binoculars}
The detailed performance (\%) of Binoculars on poems generated by different LLMs and human-written poems.
}
\end{table}

\paragraph{LLM-based Plain-text Detectors} 

Tables \ref{Qwen text}, \ref{GPT-5 text}, and \ref{Gemini text} respectively present the detailed performance (\%) of LLM-based plain-text detectors (Qwen, GPT-5, Gemini) in distinguishing between poems generated by different LLMs and human-written poems, including F1\_AI, F1\_Human, and Macro-F1.

\begin{table*}
\centering
\small
\begin{tabularx}{\textwidth}{l *{6}{>{\centering\arraybackslash}X}}
\toprule
\textbf{F1-score $\rightarrow$} & \textbf{F1\_AI} & \textbf{F1\_Human} & \textbf{Macro-F1} & \textbf{F1\_AI} & \textbf{F1\_Human} & \textbf{Macro-F1} \\
\cmidrule(lr){2-4} \cmidrule(lr){5-7}  
\textbf{Generator $\rightarrow$}  & \multicolumn{3}{c}{\textbf{GPT-4.1}} & 
 \multicolumn{3}{c}{\textbf{GLM-4}}\\
\cmidrule(lr){2-4} \cmidrule(lr){5-7} 
\TP{1} &  \textit{0.50}	& 66.39	& 33.44	& \textit{2.93}	& 66.67	& 34.80
 \\
\TP{2} &  20.69	& 67.61	& 44.15	& 27.29	& 67.81	& 47.55
 \\
\midrule
\textbf{Generator $\rightarrow$}  & \multicolumn{3}{c}{\textbf{DeepSeek-V3.1}} & 
\multicolumn{3}{c}{\textbf{DeepSeek-R1}}\\
\cmidrule(lr){2-4} \cmidrule(lr){5-7} 
\TP{1} &  \textit{0.00}	& 66.33	& 33.16	& 0.50	& 66.39	& 33.44
 \\
\TP{2} & 3.66	& 63.80	& 33.73	& 18.26	& 67.02	& 42.64
 \\
\bottomrule
\end{tabularx}
\caption{\label{Qwen text} The detailed performance (\%) of LLM-based plain-text detectors (Qwen) in distinguishing between poems generated by different LLMs and human-written poems, including F1\_AI, F1\_Human, and Macro-F1.} 
\end{table*}

\begin{table*}
\centering
\small
\begin{tabularx}{\textwidth}{l *{6}{>{\centering\arraybackslash}X}}
\toprule
\textbf{F1-score $\rightarrow$} & \textbf{F1\_AI} & \textbf{F1\_Human} & \textbf{Macro-F1} & \textbf{F1\_AI} & \textbf{F1\_Human} & \textbf{Macro-F1} \\
\cmidrule(lr){2-4} \cmidrule(lr){5-7}  
\textbf{Generator $\rightarrow$}  & \multicolumn{3}{c}{\textbf{GPT-4.1}} & 
 \multicolumn{3}{c}{\textbf{GLM-4}}\\
\cmidrule(lr){2-4} \cmidrule(lr){5-7} 
\TP{1} &  72.21	& 76.44	& 74.33	& 66.30	& 72.15	& 69.23
 \\
\TP{2} & 77.27	& 72.22	& 74.75	& 74.73	& 66.67	& 70.70
 \\
\midrule
\textbf{Generator $\rightarrow$}  & \multicolumn{3}{c}{\textbf{DeepSeek-V3.1}} & 
\multicolumn{3}{c}{\textbf{DeepSeek-R1}}\\
\cmidrule(lr){2-4} \cmidrule(lr){5-7} 
\TP{1} &  29.14	& 62.99	& 46.07	& 32.50	& 63.65	& 48.08
 \\
\TP{2} & 46.26	& 52.82	& 49.54	& 78.50	& 75.27	& 76.89 
 \\
\bottomrule
\end{tabularx}
\caption{\label{GPT-5 text} The detailed performance (\%) of LLM-based plain-text detectors (GPT-5) in distinguishing between poems generated by different LLMs and human-written poems, including F1\_AI, F1\_Human, and Macro-F1.}
\end{table*}

\begin{table*}
\centering
\small
\begin{tabularx}{\textwidth}{l *{6}{>{\centering\arraybackslash}X}}
\toprule
\textbf{F1-score $\rightarrow$} & \textbf{F1\_AI} & \textbf{F1\_Human} & \textbf{Macro-F1} & \textbf{F1\_AI} & \textbf{F1\_Human} & \textbf{Macro-F1} \\
\cmidrule(lr){2-4} \cmidrule(lr){5-7}  
\textbf{Generator $\rightarrow$}  & \multicolumn{3}{c}{\textbf{GPT-4.1}} & 
 \multicolumn{3}{c}{\textbf{GLM-4}}\\
\cmidrule(lr){2-4} \cmidrule(lr){5-7} 
\TP{1} &  80.63	& 82.52	& 81.58	& 72.03	& 77.80	& 74.92
 \\
\TP{2} & 86.49	& 87.24	& 86.86	& 74.93	& 80.00	& 77.46
 \\
\midrule
\textbf{Generator $\rightarrow$}  & \multicolumn{3}{c}{\textbf{DeepSeek-V3.1}} & 
\multicolumn{3}{c}{\textbf{DeepSeek-R1}}\\
\cmidrule(lr){2-4} \cmidrule(lr){5-7} 
\TP{1} &  64.78	& 74.62	& 69.70	& 42.16	& 67.64	& 54.90
 \\
\TP{2} & 63.94	& 74.68	& 69.31	& 83.70	& 84.99	& 84.35 
 \\
\bottomrule
\end{tabularx}
\caption{\label{Gemini text} The detailed performance (\%) of LLM-based plain-text detectors (Gemini) in distinguishing between poems generated by different LLMs and human-written poems, including F1\_AI, F1\_Human, and Macro-F1.}
\end{table*}

\paragraph{LLM-based Detectors Applying Our Method.} Tables \ref{Gemini image} and \ref{Qwen image} respectively present the  detailed performance (\%) of LLM-based detector (Gemini and Qwen) applying our proposed method on poems generated by different LLMs and human-written poems, including F1\_AI, F1\_Human, and Macro-F1.

\begin{table*}
\centering
\small
\begin{tabularx}{\textwidth}{l *{6}{>{\centering\arraybackslash}X}}
\toprule
\textbf{F1-score $\rightarrow$} & \textbf{F1\_AI} & \textbf{F1\_Human} & \textbf{Macro-F1} & \textbf{F1\_AI} & \textbf{F1\_Human} & \textbf{Macro-F1} \\
\cmidrule(lr){2-4} \cmidrule(lr){5-7}  
\textbf{Generator $\rightarrow$}  & \multicolumn{3}{c}{\textbf{GPT-4.1}} & 
 \multicolumn{3}{c}{\textbf{GLM-4}}\\
\cmidrule(lr){2-4} \cmidrule(lr){5-7} 
\IP{3}  & 63.43	& 63.07	& 63.25	& 62.93	& 61.03	& 61.98
 \\
\midrule
\textbf{Generator $\rightarrow$}  & \multicolumn{3}{c}{\textbf{DeepSeek-V3.1}} & 
\multicolumn{3}{c}{\textbf{DeepSeek-R1}}\\
\cmidrule(lr){2-4} \cmidrule(lr){5-7} 
\IP{3}  &  38.17	& 54.76	& 46.46	& 76.99	& 72.93	& 74.96
 \\
\bottomrule
\end{tabularx}
\caption{\label{Qwen image} The detailed performance (\%) of LLM-based detectors (Qwen) applying our proposed method on poems generated by different LLMs and human-written poems, including F1\_AI, F1\_Human, and Macro-F1.
}
\end{table*}

\end{document}